\documentclass[journal]{IEEEtran}
\usepackage{epsfig}
\usepackage{graphicx}
\usepackage{amsmath}
\usepackage{amssymb}
\usepackage{graphics} % for pdf, bitmapped graphics files
\usepackage{mathptmx} % assumes new font selection scheme installed
\usepackage{booktabs}
\usepackage{multirow}
\usepackage{color}
\usepackage{mathrsfs}
\usepackage{float}
\usepackage{subfigure}
\usepackage{url}
\usepackage[table]{xcolor}

\ifCLASSOPTIONcompsoc
  \usepackage[nocompress]{cite}
\else
  \usepackage{cite}
\fi

\ifCLASSINFOpdf
\else
\fi
\begin{document}
\title{Unsupervised Monocular Depth Estimation in Highly Complex Environments}

\author{Chaoqiang~Zhao,
        Yang~Tang,~\IEEEmembership{Senior Member,~IEEE,}
        Qiyu~Sun% <-this % stops a space
\thanks{\copyright 20XX IEEE. Personal use of this material is permitted.  Permission from IEEE must be obtained for all other uses, in any current or future media, including reprinting/republishing this material for advertising or promotional purposes, creating new collective works, for resale or redistribution to servers or lists, or reuse of any copyrighted component of this work in other works.}
\thanks{This work was supported in part by National Natural Science Foundation of China (Basic Science Center Program: 61988101), in part by National Natural Science Fund for Distinguished Young Scholars (61725301), in part by the Programme of Introducing Talents of Discipline to Universities (the 111 Project) under Grant B17017, in part by the Program of Shanghai Academic Research Leader (20XD1401300), in part by Innovation Research Funding of China National Petroleum Corporation (2021D002-0902), and in part by Shanghai AI Lab. (\textit{Corresponding author: Yang Tang.})}
\thanks{C. Zhao, Y. Tang, and Q. Sun are with the Key Laboratory of Smart Manufacturing in Energy Chemical Process, Ministry of Education, East China University of Science and Technology, Shanghai, 200237, China (e-mail: zhaocqilc@gmail.com, yangtang@ecust.edu.cn, qysun291@163.com).}% <-this % stops a space
}


\markboth{This work was accepted for publication in the IEEE Transactions on Emerging Topics in Computational Intelligence.}
{Shell \MakeLowercase{\textit{et al.}}: Bare Demo of IEEEtran.cls for IEEE Journals}



\maketitle

\begin{abstract}
   With the development of computational intelligence algorithms, unsupervised monocular depth and pose estimation framework, which is driven by warped photometric consistency, has shown great performance in the day-time scenario. While in some challenging environments, like night and rainy night, the essential photometric consistency hypothesis is untenable because of the complex lighting and reflection, so that the above unsupervised framework cannot be directly applied to these complex scenarios. In this paper, we investigate the problem of unsupervised monocular depth estimation in highly complex scenarios and address this challenging problem by adopting an image transfer-based domain adaptation framework. We adapt the depth model trained on day-time scenarios to be applicable to night-time scenarios, and constraints on both feature space and output space promote the framework to learn the key features for depth decoding. Meanwhile, we further tackle the effects of unstable image transfer quality on domain adaptation, and an image adaptation approach is proposed to evaluate the quality of transferred images and re-weight the corresponding losses, so as to improve the performance of the adapted depth model.
   Extensive experiments show the effectiveness of the proposed unsupervised framework in estimating the dense depth map from highly complex images. %Codes will be available.
\end{abstract}


% \fi
\IEEEpeerreviewmaketitle

\section{Introduction}

Depth is one of the most important information for autonomous systems in perceiving their surroundings and their own states \cite{zhou2021tmfnet,cadena2016past}. Therefore, the accurate estimation of the depth information from monocular images has become a hot topic in recent years and been used to improve other perception tasks \cite{sharma2020estimating}. Structure from motion and stereo matching are two main ways to recover the depth information based on the geometric relationship between images, and these methods are widely used in traditional SLAM methods to map the environments \cite{hartley2003multiple,mur2017orb}. With the development of computational intelligence algorithms \cite{gupta2017insights,li2017multiview,zhang2021survey}, deep learning algorithms have shown great performance in many tasks \cite{li2017locality,sahoo2020har}, and using deep neural networks to estimate the pixel-level dense depth from only a single image is becoming possible and has attracted much attention \cite{zhao2020monocular}. Recently, different kinds of deep learning-based monocular depth estimation frameworks have been proposed, including supervised methods, semi-supervised and unsupervised methods \cite{eigen2014depth,zhang2021multitask,zhao2020masked}. Because of the costly ground truth, geometric constraints are gradually replacing ground truth for the training of depth networks, and the unsupervised framework has become a promising direction for monocular depth estimation \cite{poggi2020uncertainty}.

\begin{figure}[t]
	\centering
	\includegraphics[width = \columnwidth]{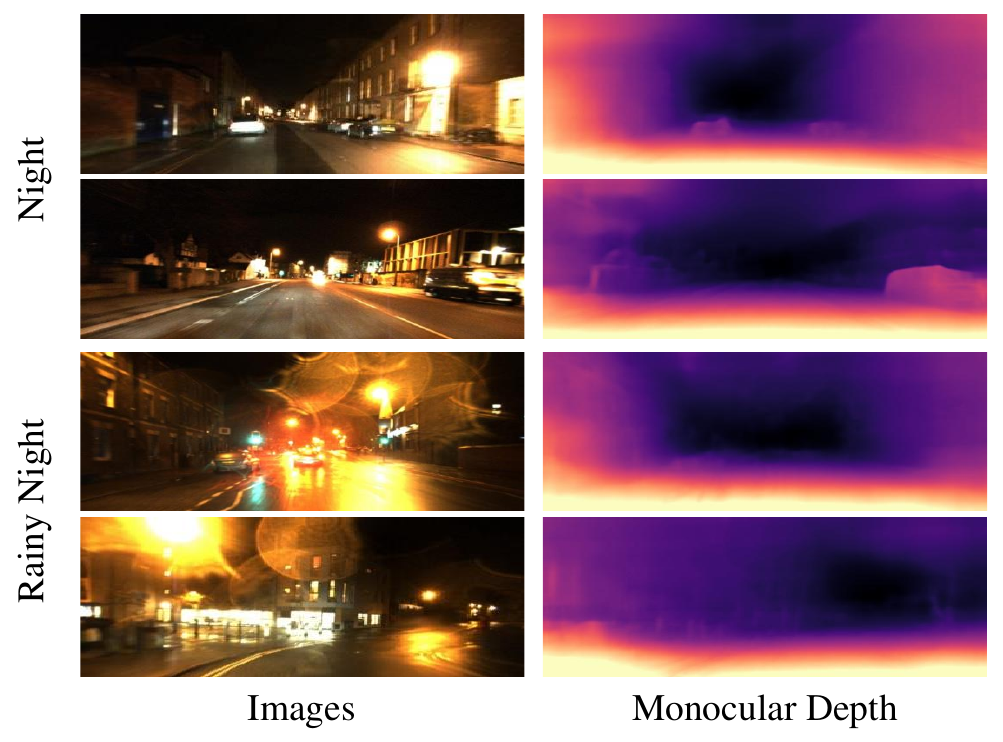}
	\caption{The monocular depth predictions of the proposed method at night and rainy night. Note that the encoders of the above three depth models are adapted from the same basic encoder, and their decoders are the same.}
	\label{fig:fig1}
\end{figure}

In the unsupervised framework of monocular depth estimation \cite{zhou2017unsupervised,sun2021unsupervised}, the geometric constraints between adjacent images are considered to supervise the network training. Therefore, only monocular image sequences and camera parameters are needed during the training process. This unsupervised framework is mainly composed of two deep neural networks, including a depth network to regress the dense depth from single images and a pose network to estimate the pose between two frames. Based on the estimated depth map and pose, the geometric relationship between images is built on the projection function. The mainly supervised signal is calculated from the photometric error of corresponding pixels between adjacent images by using view reconstruction \cite{zhou2017unsupervised,godard2019digging}.

However, since the unsupervised signal is built on projection consistency, the above unsupervised framework \cite{godard2019digging,zhou2017unsupervised} suffers from two big limitations, \textit{the static scenario hypothesis} and \textit{the photometric consistency hypothesis}. For \textit{the static scenario hypothesis}, since the pixels on moving objects do not satisfy the projection function of camera ego-motion, which leads to incorrect calculation of the loss during training, thereby affecting the accuracy of the depth network \cite{lee2021learning}. Incorporating semantic information into the unsupervised framework is an effective way to recognize moving objects and eliminate their influence, and relevant results have emerged \cite{klingner2020self,lee2021learning}.
For \textit{the photometric consistency hypothesis}, since the training of the unsupervised framework relies heavily on the photometric error, the photometric consistency of the same pixel on different images is crucial to the overall framework \cite{spencer2020defeat}. Since all objects are illuminated by the same light source (sun) during the day, the photometric consistency assumption is basically valid, just like that in traditional direct visual odometer methods \cite{engel2017direct,yang2018challenges}. Therefore, almost all of the current unsupervised monocular depth estimation methods are trained and tested on day-time images.
When the environment changes into other highly complex conditions, like \textit{night} and especially \textit{rainy night}, static objects with non-Lambertian or high reflectance surfaces in the night/rainy night with dynamic lighting conditions violate the photometric consistency between frames. Besides, objects with low luminance in the night lack reliable cues for providing accurate correspondences.
The problem of estimating monocular depth in such varying environments is challenging but practical and important. Meanwhile, the perception of changing and complex environments is crucial for autonomous systems \cite{tang2020overview,tang2020perception}, like robots and autonomous driving cars, and this problem has only received some initial attention \cite{spencer2020defeat,vankadari2020unsupervised}.

Because of the limitations of the unsupervised framework in complex scenarios, we tackle this challenging problem by using an image transfer based domain adaptation framework. Instead of training the unsupervised framework on the images from complex environments, we adapt the model trained by day-time images to other complex environments, thereby circumventing the photometric inconsistency in complex environments and achieving satisfactory depth estimation in an unsupervised way in the highly complex environments.
We only adapt the encoder of the depth network by following \cite{vankadari2020unsupervised}. An additional encoder is designed to encode the images of complex environments, and after adaptation, the encoders for complex environments share the same feature space with the day-time encoder. Besides, the adapted encoders for different scenarios share the same decoder for monocular depth estimation in different complex scenarios, which is meaningful for practical applications. Therefore, this method not only reduces computational complexity but also facilitates practical applications: switching different encoders for adapting various environments.
Different from \cite{vankadari2020unsupervised} using adversarial domain adaptation in \textit{feature space} for night-time depth estimation, we propose to adopt the image transfer-based domain adaptation framework and constrain the training from both \textit{feature space} and \textit{output space}, which is more stable and accurate \cite{chen2019crdoco,kim2019diversify}. %\cite{chen2019crdoco,hoffman2018cycada,kim2019diversify,zhang2018fully}. %and have been widely used in related tasks for domain adaptation . %Besides, we also propose to constrain the consistency of information passed by skip connection between encoder and decoder, which is important but ignored by \cite{vankadari2020unsupervised}.
Besides, for image transfer based domain adaptation framework, the generated paired images are used to get pseudo labels from the known models and supervise the adaptation process. Poor generated images result in wrong pseudo labels and affect the training process, which is not considered in previous works.
Therefore, in this paper, we consider the errors introduced by the unstable image transfer, and an image quality adaptation approach is proposed to evaluate the quality of the transferred images and reduce their effects.
The proposed image transfer-based domain feature adaptation (ITDFA) framework not only can be used for night-time depth estimation, but also shows outstanding performance on more challenging rainy night-time images, as shown in Fig. \ref{fig:fig1}.% Moreover, the models obtained by ITDFA can extend to adapt to new scenarios, which demonstrates the ability of ITDFA to learn and extract the key features for depth decoding.

%The pre-trained CycleGAN model is used to convert the day-time and the complex scenario images to each other, and the day-time encoded features are used as the reference to align with the trained encoder.

In summary, in this paper, we analyze and tackle the unsupervised monocular depth estimation problem in three typical and challenging scenarios (night, rainy night), including proposing the ITDFA framework, constructing novel training/testing sets on different scenarios, digging into the continuous adaptation ability of the ITDFA, and exploring the influence of the image transfer model on ITDFA.
Our main contributions are as follows:
\begin{itemize}
\item This paper analyzes the major reason for the limited performance and application of the current unsupervised monocular depth estimation framework, and we tackle the problem of unsupervised monocular depth estimation in highly complex environments by using domain adaptation.
\item Image transfer-based unsupervised domain adaptation is applied to estimate monocular depth from challenging scenarios, like night and rainy night. To reduce the effects of the unstable image transfer quality, we propose an image quality adaptation approach to evaluate the quality of the transferred images and re-weight the corresponding losses.
\item Extensive experiments and results on the RobotCar dataset \cite{RobotCarDatasetIJRR} show the effectiveness of our proposed method in highly complex environments.
\end{itemize}

%Our paper is organized as follows: first,

%------------------------------------------------------------------------
\section{Related work}

In this section, we introduce the popular unsupervised monocular depth estimation framework \cite{zhou2017unsupervised}, which is trained on monocular sequences. Firstly, many recent research results for improving this unsupervised framework are briefly reviewed, from the perspectives of occlusions, static scenario hypothesis and photometric consistency hypothesis. Then, we review the framework combined with domain adaptation, in which depth models are trained on synthetic datasets and then adapted to real-world scenarios through domain adaptation.
%------------------------------------------------------------------------

\textbf{Unsupervised framework.}
To circumvent the need for costly ground truth, Zhou \textit{et al.} \cite{zhou2017unsupervised} propose to use geometric constraints between frames instead of ground truth to train a depth network. Their framework contains a depth network for monocular depth estimation and a pose network for inter-frame pose estimation. Then, based on the projection function established by the estimated pose and depth, the view reconstruction is designed to warp and construct the target frame from its adjacent frame. The photometric error between the warped and real target images is used to supervise the training process, so that the depth and pose networks are trained in an unsupervised manner. To improve the accuracy of depth estimation, several novel loss functions and network frameworks are proposed, which are well reviewed in \cite{zhao2020monocular}.

\begin{figure*}[t]
	\centering
	\includegraphics[width = 1.8\columnwidth]{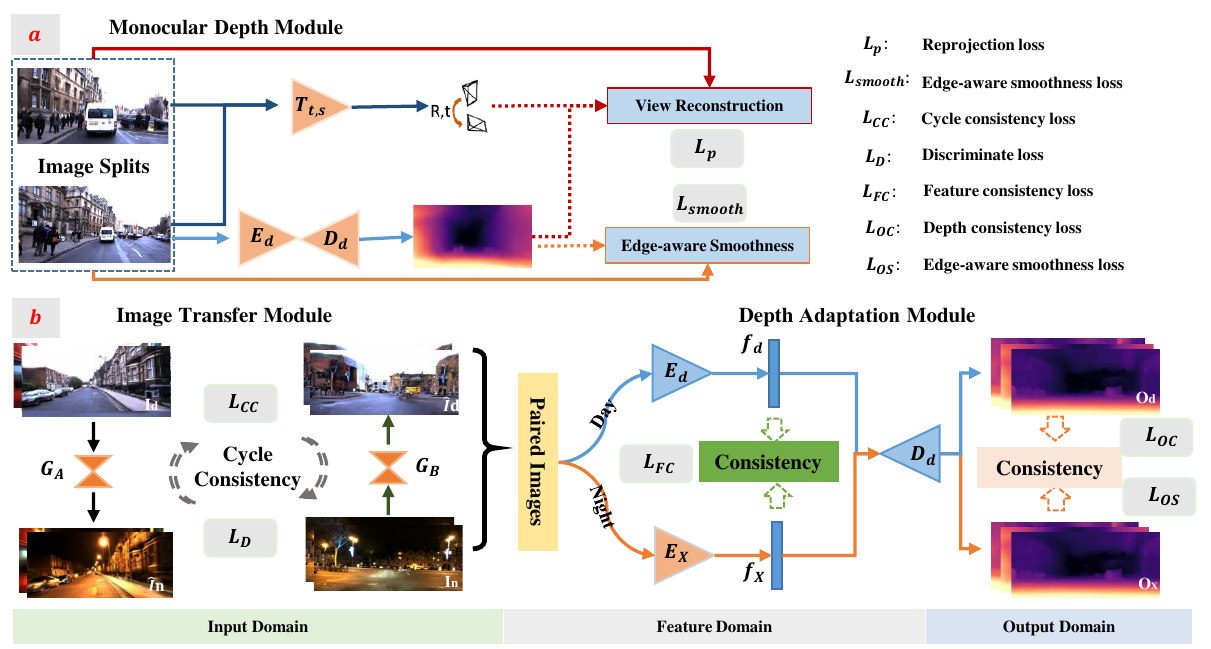}
	\caption{The framework of ITDFA for unsupervised monocular depth estimation in highly complex environments. In (a), the unsupervised monocular depth estimation framework follows monodepth2 \cite{godard2019digging}. In (b), our ITDFA framework contains two modules, image transfer module and depth adaptation module. $I_{d}$ and $I_{X}$ refer to the image from day-time scenario and one of the complex scenarios, and $O_{d}$ and $O_{X}$ stand for their corresponding depth maps. The encoder $E_{d}$ and decoder $D_{d}$ trained by day-time images do not update their weights during training.}
	\label{fig:fig2}
\end{figure*}

\textbf{Photometric inconsistency.} Photometric inconsistency is one of the major reasons for the limited performance and application of the unsupervised framework.
%Similar to the drawbacks faced by direct visual odometry \cite{von2020gn}, the above unsupervised framework is also limited by the photometric consistency hypothesis. Affected by the camera optical vignetting and exposure controls, the corresponding pixels between day-time images face the problem of photometric inconsistency \cite{yang2018challenges}, which results in the performance degradation of the unsupervised framework in outdoor scenarios. To deal with the photometric inconsistency, Yang \textit{et al.} \cite{yang2020d3vo} propose to estimate the brightness transformation parameters to align the photometry of images during training.
For some highly complex environments, like \textit{night-time environments} and more challenging \textit{rainy night-time environments}, due to the complex lighting conditions, e.g., street lamps, car lights, and especially the reflection of light from the road caused by rain, the essential photometric consistency hypothesis is untenable, so the unsupervised framework shows unsatisfied accuracy and robustness.
The unsupervised monocular depth estimation in such complex scenarios is a largely under explored domain, and only a few methods for night-time monocular depth estimation have been proposed most recently  \cite{spencer2020defeat,vankadari2020unsupervised}.
To overcome the photometric consistency of images, Spencer et al. \cite{spencer2020defeat} propose to use dense feature representation of images for unsupervised training.
%A deep neural network is proposed to extract the feature maps of images, which are consistent under different lighting conditions.
Since the corresponding features between different images are consistent and unaffected by light, this unsupervised framework can well adapt to night-time scenarios. Nevertheless, since the whole framework is unsupervised, their framework still needs the help of photometric error during training. Different from \cite{spencer2020defeat}, Vankadari \textit{et al.} \cite{vankadari2020unsupervised} regard this challenge as a domain adaptation problem. The depth and pose networks are trained on day-time scenarios by following \cite{godard2019digging} at first. Then, an additional encoder is designed to encode the night-time images, and an adversarial domain feature adaptation method is used to adapt the features encoded by the day-time encoder and night-time encoder. %Finally, the combination of night-time encoder and day-time decoder is used for night-time monocular depth estimation.
Since the output of the encoder are multi-scale high-dimensional feature maps, they design multiple discriminators to constrain each scale feature map.
However, adjusting the adversarial framework consisting of multiple discriminators and a generator is extremely complex, which influences its stability for applying to other scenarios \cite{salimans2016improved,wang2017generative}. Moreover, although adversarial learning helps to reduce the distance between the distributions of day and night feature spaces, the key features for depth decoding are not valued because the decoder is not involved in their domain adaptation process.

\textbf{Domain adaptation.}
%这一块需要再重新好好写写
Due to the domain shift, like differences in the background, lighting, weather, and so on of images between different datasets/domains, the performance of the trained model may degrade significantly when it was applied to other datasets \cite{tommasi2017deeper,ye2019deep,9565825}. Therefore, a domain adaptation framework is proposed to transfer the model from one domain to another for the same task \cite{wilson2020survey}, and most recently, adaptation between multiple domains has received a lot of attention \cite{roy2021curriculum,ahmed2021unsupervised}.

In monocular depth estimation, domain adaptation algorithms are mainly applied to adapt the model trained on synthetic datasets to real-world datasets \cite{kundu2018adadepth,gu2020coupled,zhao2019geometry}. Compared with the ground truth obtained by different sensors in the real-world, the ground truth obtained from virtual environments is cheaper and easier. The depth model is trained on synthetic and real-world datasets and supervised by the ground truth of synthetic datasets. Since supervised training can get more cues than unsupervised training, this method achieves better accuracy on monocular depth estimation than unsupervised methods, and it provides a new way to circumvent the need for costly ground truth at the same time. Most recently, the LAB-based images transfer approach is proposed to transfer images between domains for domain adaptation \cite{he2021multi}. While for the night-time scenario, the above frameworks cannot work because it is difficult to generate synthetic night-times images that can capture all the vagaries of real-world night conditions \cite{vankadari2020unsupervised}, let alone rainy night-time scenarios and even more complex scenarios. Therefore, to tackle the unsupervised monocular depth estimation in highly complex environments, we use unsupervised domain adaptation to transfer the model trained on day-time images to work for night-time and rainy night-time images.% To increase the applicability and reduce the computational complexity, our model only adapts the encoders during training by following \cite{vankadari2020unsupervised}. Similar to the concurrent work \cite{liu2021self}, CycleGAN-based image transfer method is introduced to generate paired images between day and night because it can well mimics nighttime lighting conditions. Different with

To increase the applicability and reduce the computational complexity, our model only adapts the encoders during training by following \cite{vankadari2020unsupervised}, but our model ITDFA does not need to consider the stability of adversarial learning. Meanwhile, Vankadari \textit{et al.} \cite{vankadari2020unsupervised} only consider the adversarial constraint on feature space, while this paper constrains the training of the encoder from both feature space and output space. The constraints on output space help the encoder to focus on learning the key features of depth decoding. Similar to the concurrent work \cite{liu2021self}, the CycleGAN-based image transfer method is introduced to generate paired images between day and night because it can well mimic nighttime lighting conditions, and the transferred images are used to generate pseudo labels and then supervise the training process. However, as shown in Fig. \ref{fig:fig3} (a), poor transferred images will generate wrong pseudo labels and supervised signals, which will affect the performance of the network.
We want to tackle the above problems by evaluating the transferred images and thus reducing their effect on training. However, since the transfer models are trained in an unsupervised manner, and there are no real paired images between domains, it is difficult to demonstrate the quality of transferred images.
After lots of experiments, we find a new method to reflect the quality of the transferred images, which is described in the next section in detail. Hence, our proposed image transferred domain adaptation model achieves a better performance than \cite{vankadari2020unsupervised,liu2021self} in the night-time scenario. Moreover, they \cite{vankadari2020unsupervised,liu2021self} only address the unsupervised monocular depth estimation on night-time images, while our framework can also do well in more challenging rainy night-time scenario.

\textbf{Discussion:}
\textit{Why not adopt a framework that directly combines image style transfer with monocular depth estimation:} the images from complex environments are first transferred to normal day-time style and then use the day-time depth model to estimate the depth. After testing, we find this is a possible way to solve the problem, as shown in lines 1-2 of Fig. \ref{fig:fig3} (a). Nevertheless, poor real-time performance will limit the application of this framework because of the two-step process. Besides, since the accuracy of depth estimation relies heavily on the quality of transferred images, this approach has great instability in practical applications, as shown in lines 3-4 of Fig. \ref{fig:fig3} (a). The proposed ITDFA framework can get a new model for the new scene and predict the depth in an end-to-end manner. Moreover, since only the encoder is trained, our method can get multiple encoders for multiple scenes, and these encoders share the same decoder, which is very practical.

%\textcolor{blue}{Image transfer based unsupervised domain adaptation methods are proposed to train a new and independent depth model for new scene. Since the pseudo labels generated by the transferred images }
%Evaluating the transferred images and thus reducing their effect on training is the effective way to tackle the above problems. However, since the transfer models are trained in an unsupervised manner, and there are no paired images between domains, it is difficult to demonstrate the quality of transferred images.

\section{Methods}
In this section, we will introduce the overall ITDFA framework, loss functions for training as well as the image quality adaptation strategy proposed in this paper.

\subsection{Unsupervised depth estimation part}

 We use the famous unsupervised framework, monodepth2 \cite{godard2019digging}, to acquire the trained depth network in an unsupervised manner, which is shown in Fig. 2-a. Monodepth2 \cite{godard2019digging} has been widely used as the basic unsupervised framework in this field because of its high practicability and accuracy \cite{johnston2020self,klingner2020self,spencer2020defeat}. For the day time scenario, the unsupervised monocular depth estimation is formulated as the minimization of the per-pixel minimum reprojection error:
\begin{equation}
\mathrm{L}_{p} = min \Psi (I_{t},I_{s \to t}),  \label{eq:0}
\end{equation}
\begin{equation}
\Psi(I_{a},I_{b}) = \dfrac{\alpha}{2} (1-SSIM(I_{a},I_{b})) + (1-\alpha) ||I_{a}-I_{b}||_{1}, \label{eq:00}
\end{equation}
and
\begin{equation}
I_{s \to t} = I_{s} \langle proj(O_{t},T_{t \to s},K) \rangle,  \label{eq:000}
\end{equation}
where $||.||_{1}$ refers to the L1 distance in pixel space, and $proj$ stands for the 2D coordinate projection based on the predicted dense depth map $O_{t}$ of the target image $I_{t}$ and the related pose $T_{t \to s}$ between target $I_{t}$ and source images $I_{s}$. Meanwhile, the edge-aware smoothness loss is also used to improve the depth map $O_{t}$:
\begin{equation}
\mathrm{L}_{smooth} = |\partial_{x} o_{t}^{*}|e^{\partial_{x} I_{t}} + |\partial_{y} o_{t}^{*}|e^{\partial_{y} I_{t}}, \label{eq:0000}
\end{equation}
where $d_{t}^{*}=o_{t}/\hat{o_{t}}$ represents the mean-normalized inverse depth.

The monodepth2 framework is supervised by combining the per-pixel smoothness loss and masked photometric loss on the day-time scenario. The depth model consisting of an encoder $E_{d}$ and a decoder $D_{d}$ is used in the proposed ITDFA. The depth model learns a mapping from the day-time images $I_{d}$ to the pixel-level depth maps $O_{d}$:
\begin{equation}
O_{d} =  D_{d} (E_{d}(I_{d})).  \label{eq:1}
\end{equation}

\subsection{Image transfer part}

Since the LAB-based image transfer approach \cite{he2021multi} cannot well simulate the complex and heterogeneous lighting conditions at night, we utilize the CycleGAN-based framework \cite{CycleGAN2017} to transfer the images between scenarios. During training, the full objective is:
\begin{equation}
G_{d2X},G_{X2d} = arg \min_{G} \max_{D} \mathrm{L}(G_{d2X},G_{X2d},D_d,D_X) , \label{eq:001}
\end{equation}
\begin{equation}
\mathrm{L}(G_{d2X},G_{X2d},D_d,D_X) = \mathrm{L_{CC}} + \mathrm{L_{D}} , \label{eq:01}
\end{equation}
where $G_{d2X}$ tries to generate images $G_{d2X}(I_{d})$ that look similar to images from domain $X$, while $D_X$ aims to distinguish between translated samples $G_{d2X}(I_{d})$ and real samples $X$. $L_{CC}$ and $L_{D}$ refer to the cycle consistency loss and adversarial loss:
\begin{equation}
\begin{array}{l}
    \mathrm{L}_{CC}(G_{d2X},G_{X2d}) = \mathbb{E}_{I_{d} \sim p_{data}(I_{d})}[||G_{X2d}(G_{d2X}(I_{d}))-I_{d}||_{1}]  \\
    \qquad  \qquad  \qquad   + \mathbb{E}_{ I_{X} \sim p_{data}(I_{X})}[||G_{d2X}(G_{X2d}(I_{X}))-I_{X}||_{1}]
\end{array}  ,\label{eq:002}
\end{equation}
and
\begin{equation}
    \mathrm{L}_{D} = \mathrm{L_D}(G_{d2X}, D_X, I_{d}, I_{X}) + \mathrm{L_D}(G_{X2d}, D_d, I_{X}, I_{d}) ,\label{eq:003}
\end{equation}
and all of the above constraints used for training image transfer models ($G_{d2X}$ and $G_{X2d}$) are following Zhu \textit{et al.} \cite{CycleGAN2017}.

In this paper, for different complex environments, different image transfer models ($G_{d2X}$ and $G_{X2d}$) based on CycleGAN \cite{CycleGAN2017} are needed to transfer the images between day-time style $d$ and different complex environmental styles $X$:
\begin{equation}
\left\{ \begin{array}{ll}
\hat{I}_{d2X} = G_{d2X}(I_{d}), & \textrm{day to X}\\
\hat{I}_{X2d} =  G_{X2d}(I_{X}), & \textrm{X to day}\\
\hat{I}_{d2X2d} = G_{X2d}(G_{d2X}(I_{d})), & \textrm{cycle transfer}\\
\hat{I}_{X2d2X} =  G_{d2X}(G_{X2d}(I_{X})), & \textrm{cycle transfer}
\end{array} , \right. \label{eq:2}
%\end{array}{1}
\end{equation}
where $X$ refers to night $n$ or rainy night $r$. In addition, to verify the continuous transfer ability of the model obtained through domain adaptation, we also train an additional image transfer model between night-time style and rainy night-time style ($G_{n2r}$ and $G_{r2n}$).

%To reduce the complexity of the whole framework, we donot
%In this paper, we focus on the unsupervised monocular depth estimation in night-time, rainy night-time and snowy winter scenarios. Therefore, there are three image transfer models are pre-trained: between day-time style and night-time style ($G_{d2n}$ and $G_{n2d}$), between day-time style and rainy night-time style ($G_{d2rn}$ and $G_{rn2d}$), and between day-time style and snowy winter style ($G_{d2sw}$ and $G_{sw2d}$). In addition, to verify the continuous transfer ability of the model obtained through domain adaptation, we also trained an additional image transfer model between night-time style and rainy night-time style ($G_{n2rn}$ and $G_{rn2n}$).
%The effectiveness of the proposed system depends on the quality of image transfer.

\subsection{ITDFA part}

As shown in Fig. 2-c, an encoder $E_{X}$ is designed to encode the features of images from highly complex scenarios to the same feature space as the features of day-time images encoded by the day-time encoder $E_{d}$. In ITDFA, the pre-trained day-time encoder $E_{d}$ is used to encode the day-time images and obtain their corresponding feature maps $f$:
 \begin{equation}
\left\{ \begin{array}{l}
f_{d} = E_{d}(I_{d})\\
f_{X2d} =  E_{d}(\hat{I}_{X2d})
\end{array} , \right. \label{eq:3}
%\end{array}{1}
\end{equation}
%where $I_{d}$ and $\hat{I}_{X2d}$ refer to the real and transferred day-time images.
where $I_{d}$ refers to the real day-time images, and $\hat{I}_{X2d}$ stands for the fake day-time images generated by CycleGAN model $G_{X2d}$ from the highly complex scenario $X$.
The encoder $E_{X}$ for complex environments has the same network framework as the day-time encoder $E_{d}$, and it is used to encode the real and fake images of highly complex scenarios and obtain their feature maps $f$:
 \begin{equation}
\left\{ \begin{array}{l}
f_{X} = E_{X}(I_{X})\\
f_{d2X} =  E_{X}(\hat{I}_{d2X})
\end{array} , \right. \label{eq:4}
%\end{array}{1}
\end{equation}
where $I_{X}$ refers to the real images from highly complex scenarios, and $\hat{I}_{d2X}$ stands for the fake images transferred from day-time scenario $d$.
The pre-trained decoder $D_{d}$ is used to decode the features from $E_{d}$ and $E_{X}$ and obtain their corresponding depth maps $O$:
 \begin{equation}
\left\{ \begin{array}{l}
O_{d} = D_{d}(f_{d})\\
O_{X2d} =  D_{d}(f_{X2d})\\
O_{X} = D_{d}(f_{X})\\
O_{d2X} =  D_{d}(f_{d2X})
\end{array} . \right. \label{eq:5}
%\end{array}{1}
\end{equation}
During training, the weights of $E_{d}$ and $D_{d}$ are fixed, and only $E_{X}$ is updated.

During testing, the depth map can be estimated from the images of highly complex scenarios in one-step:
\begin{equation}
O_{X} =  D_{d} (E_{X}(I_{X})).  \label{eq:6}
\end{equation}

\begin{figure*}[t]
	\centering
	\subfigure[Errors introduced by unstable image transfer]{
		\includegraphics[width = 0.85\columnwidth]{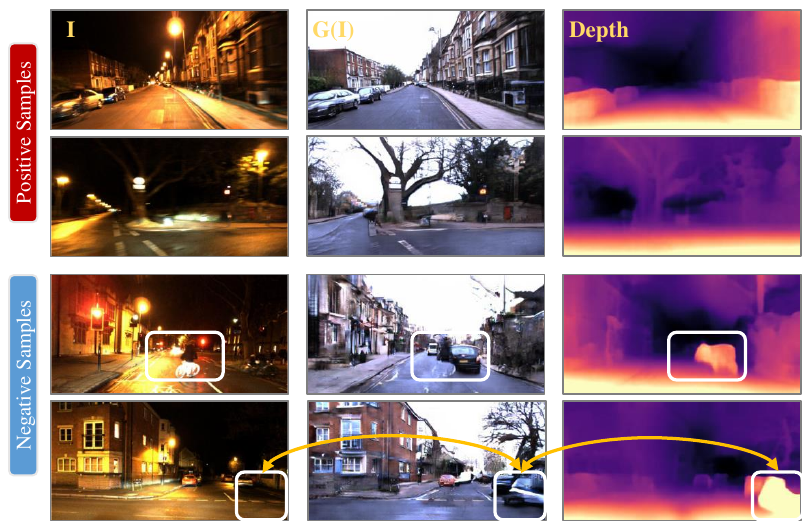}
		%\caption{fig1}
	}
	\subfigure[Methods to reflect the quality of transferred images]{
		\includegraphics[width = 1.05\columnwidth]{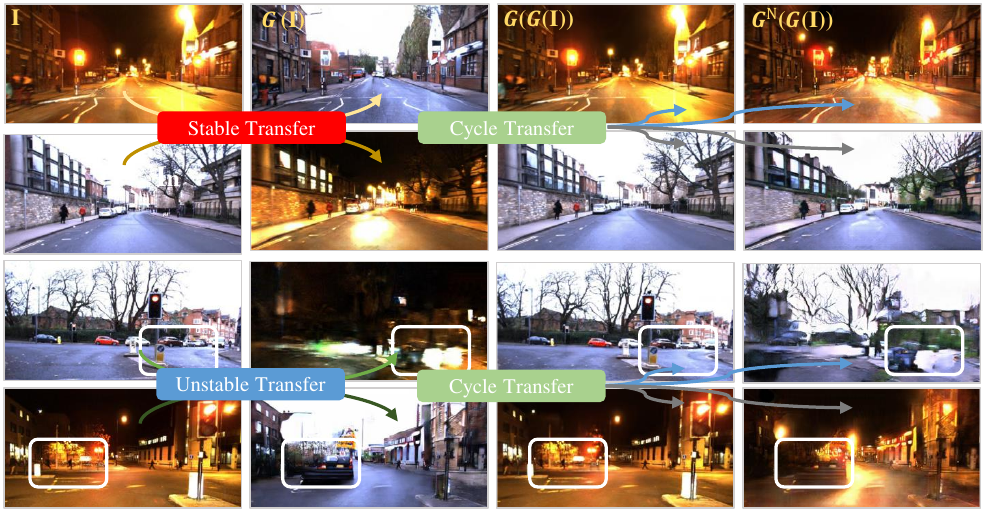}
	}
	\caption{Samples of applying CycleGAN \cite{CycleGAN2017} for monocular depth estimation in complex environments. As shown in (a), since the CycleGAN-based image transfer method is instable, the wrong image conversion will introduce the incorrect depth estimation into training. To reduce the effects of unstable transfer on domain adaptation, the quality of transferred image (2nd column in (b)) should be accurately evaluated, and our proposed method reflects the quality effectively(4th column in (b)).}
	\label{fig:fig3}
\end{figure*}

\textbf{Training losses:} The ITDFA framework is an unsupervised framework, and neither ground truth nor real paired images are used to train the depth model, CycleGAN model and domain adaptation model. As shown in Fig. \ref{fig:fig2}, different constraints are designed to supervise the training process, including the feature consistency loss $\mathrm{L}_{FC}$ on feature space, and the depth consistency loss $\mathrm{L}_{OC}$ as well as smoothness loss $\mathrm{L}_{OS}$ on output (depth) space. Therefore, the overall loss function for training the encoder $E_{X}$ is formulated as:
\begin{equation}
\mathrm{L}_{DA} =  \mathrm{L}_{FC} + \beta \mathrm{L}_{OC} + \gamma \mathrm{L}_{OS}.  \label{eq:7}
\end{equation}

\textbf{Feature consistency loss:} Based on the pre-trained CycleGAN model, we can get the paired images from day-time scenario and highly complex scenario, like $I_{d}$ with $I_{d2X}$ and $I_{X}$ with $I_{X2d}$. Therefore, to promote the consistency of different encoders in feature space, we direct minimize the error of the feature maps, which are encoded by $E_{d}$ and $E_{X}$ from these image pairs:
\begin{equation}
\mathrm{L}_{FC_{L1}} = L1 (f_{d},f_{d2X})+ L1 (f_{X},f_{X2d}).  \label{eq:8}
\end{equation}
Moreover, inspired by style transfer methods and related works \cite{lee2021dranet,cheng2021style,liu2021self}, to enhance the consistency of correlations between features, the Gram Matrices $\mathscr{G}$ between features are calculated to further improve the feature consistency:
\begin{equation}
\begin{array}{ll}
\mathrm{L}_{FC} &= \mathrm{L}_{FC_{L1}} + \alpha \mathrm{L}_{FC_{Gram}} \\
                &= L1 (f_{d},f_{d2X})+ L1 (f_{X},f_{X2d}) \\
                &+ L1 (\mathscr{G}(f_{d}),\mathscr{G}(f_{d2X})) + L1 (\mathscr{G}(f_{X}),\mathscr{G}(f_{X2d}))
\end{array}.  \label{eq:88}
\end{equation}
$\alpha$, $\beta$ and $\gamma$ are the weights of each loss function for training.

\textbf{Depth consistency loss:} Although the above loss can help to promote the consistency of feature space between the two encoders, the ultimate goal is the depth map rather than the feature map, and the contribution of different features to the depth decoding is different. To further constrain the key features of feature maps for depth decoding, we design a depth consistency loss in output space:

\begin{equation}
\mathrm{L}_{OC} =  L1 (O_{d},O_{d2X}) + L1 (O_{X},O_{X2d}).  \label{eq:9}
\end{equation}

\textbf{Smoothness loss:} Moreover, to promote the smoothness of the generated depth map, we propose to utilize the edge-aware smoothness during training, which is widely used in previous unsupervised depth framework \cite{godard2017unsupervised,godard2019digging,wang2018Learning}:
\begin{equation}
\mathrm{L}_{OS} = |\partial_{x} O_{d2X}^{*}|e^{\partial_{x} I_{d}} + |\partial_{y} O_{d2X}^{*}|e^{\partial_{y} I_{d}}, \label{eq:11}
\end{equation}
where $O_{d2X}^{*}=O_{d2X}/\bar{O}_{d2X}$ represents the mean-normalized inverse depth. Note that this loss is established between the real day-time image $I_{d}$ and the depth map $O_{d2X}$ of its corresponding transferred images $\hat{I}_{d2X}$.

%\begin{figure*}[t]
%	\centering
%	\includegraphics[width = 1.8\columnwidth]{figures/night.pdf}
%	\caption{\textbf{Estimating the depth from night-time images.} ``Monodepth2(n)'' refers to the depth model trained on night-time images by the unsupervised monocular framework, monodepth2. For night-time scenario, it suffers from complex lighting conditions, like the street lights, car lights, and the reflection of light from the road, which bring big challenges to the unsupervised training framework. }
%	\label{fig:fig7}
%\end{figure*}

\subsection{Image quality adaptation part}

Inspired by the cycle consistency of CycleGAN during the training, we try to use the consistency between the raw image and cycle transferred image to report the performance. Nevertheless, this method cannot work well because of the overfitting, as shown in columns 1 and 3 of Fig. \ref{fig:fig3} (b). After a series of tests, we found that using the models saved from different epoch can demonstrate the quality of the transferred images.
As shown in column 4 of Fig. \ref{fig:fig3} (b), for the good generated images, our method can generate good cycle images; while for the poor generated images, our method can well reflect the poor regions of the images. Therefore, we compute the SSIM loss between the raw image $(I)$ and cycle transferred image $\hat{I}_{cycle}=G^{N}(G(I))$ to quantify the quality of the transferred image $\hat{I}=G(I)$. $G^{N}$ refers to a fixed transfer model saved at initial epoch $N$, in this paper, we set `N' $=$ 30 in the training process.

During training, unpaired day and night time images are sent to the transfer network and depth network during training. Since the quality of transferred images varies from model to model and image to image, we evaluate the image quality and re-weight their domain adaptation loss $L_{DA}$ based on our image adaptation method:
\begin{equation}
L_{total} = (1-\eta) L_{DA}^{d} + \eta L_{DA}^{X}, \label{eq:20}
\end{equation}
where $L_{DA}^{d}$ and $L_{DA}^{X}$ refer to the losses calculated from day-time input image flow and night/rainy night image flow. $\eta$ stands for:
 \begin{equation}
\eta =  \frac{(1-SSIM(I^{d},\hat{I}^{d}_{cycle}))}{((1-SSIM(I^{d},\hat{I}^{d}_{cycle}))+(1-SSIM(I^{X},\hat{I}^{X}_{cycle})))}. \label{eq:21}
\end{equation}

However, the above weight $\eta$ can only adjust the training process to pay more attention to good transferred image samples in each input pair, and it cannot completely eliminate the effects of poor transferred images.
In the training framework, the pseudo depth labels are the depth maps predicted by pre-trained daytime models from real and transferred day-time images. As shown in the negative samples in Fig. \ref{fig:fig3} (a), if we directly use $L1$ to constrain the consistency of depth maps (Eq. \ref{eq:9}), the night-time depth model will be supervised by wrong labels.

Therefore, to filter the errors introduced by instable image transfer, inspired by the minimization loss used in monodepth2 \cite{godard2019digging}, we propose a minimization loss to solve the above errors:
\begin{equation}
\begin{array}{ll}
\mathrm{L}_{OC} = &Min(<O_{d},O_{d2X}>,<O_{d2X},O_{d2X2d}>) \\
                &+ Min(<O_{X},O_{X2d}>,<O_{X2d},O_{X2d2X}>)
\end{array},  \label{eq:22}
\end{equation}
where $O_{d2X2d}$ and $O_{X2d2X}$ stand for the depth maps predicted by the cycle transferred images $\hat{I}^{n}_{d2X2d} = G_{X2d}(G^{n}_{d2X}(I_{d}))$ and $\hat{I}^{n}_{X2d2X} = G^{n}_{d2X}(G_{X2d}(I_{d}))$. As shown in the negative samples in Fig. \ref{fig:fig3} (b), the proposed cycle transferred method can reflect the quality of transferred images, which means that the depth of transferred images is consistent with the depth of raw image or the depth of cycle transferred images. With the help of minimization loss, $\mathrm{L}_{OC}$ helps the network learn from the more accurate pseudo labels.

\section{Experiments}

\subsection{Datasets}
Since this paper focuses on the unsupervised monocular depth estimation in multiple highly complex environments, we choose the publicly available Oxford RobotCar dataset \cite{RobotCarDatasetIJRR} as our training and testing sets. RobotCar dataset \cite{RobotCarDatasetIJRR} is one of the most famous outdoor datasets, and it contains the image sequences collected in all weather conditions, including rain, night, direct sunlight and snow. The image sequences captured by the left camera of Bumblebee XB3 are used for the experiments of this paper. The images are manipulated to RGB style from the raw recordings with the resolution of 1280x960, and we crop the car-hood of the images and resize them to 512x256. For the day-time and night-time scenarios, we use the sequences from 2014-12-09-13-21-02 and 2014-12-16-18-44-24, which are the same as \cite{vankadari2020unsupervised} for a fair comparison. For the rainy night-time scenario that have not received attention in recent research, we choose the sequences from 2014-12-17-18-18-43.

%\begin{figure*}[h]
%	\centering
%%%	\includegraphics[width = 1.8\columnwidth]{figures/rainynight.pdf}
%	\caption{\textbf{Estimating the depth from rainy night-time images.} ``Monodepth2(r)'' refers to the depth model trained on rainy night-time images by the unsupervised monocular framework, monodepth2. For the more complex rainy night scenario, it suffers from not only the complex lighting conditions at night but also the complex reflections on the road and camera caused by rain.}
%	\label{fig:fig8}
%\end{figure*}

\begin{table*}[t]
	
	\scriptsize
	
	\centering
	
	\caption{\textit{Comparison with the unsupervised depth estimation methods for night-time scenarios. ``M'' means that the supervisory signals mainly come from monocular sequences.}}
	
	\label{Tab01}
	\begin{tabular}{c|c|c|cccc|ccc}
		
		\toprule
		\multicolumn{3}{c}{}& \multicolumn{4}{c}{Error Metrics (Lower is better)} & \multicolumn{3}{c}{Accuracy (Higher is better)}  \\
		\cmidrule(r){1-3}\cmidrule(r){4-7} \cmidrule(r){8-10}
		\hline
		Method   	&  Supervision     &   Depth-range (m)		&  Abs Rel      &  Sq Rel    &   RMSE 	&  RMSE log     &   $\delta < 1.25^{1}$		&  $\delta < 1.25^{2}$      &  $\delta < 1.25^{3}$ \\
		\hline
		\rowcolor{gray!9}
		Monodepth2 (d) \cite{godard2019digging} 	&  M     &   40			&  0.4240  &  3.8665   &   8.3071	&  0.4562   &   0.317	&  0.595    & 0.828 \\
		Monodepth2 (n) \cite{godard2019digging} 	&  M     &   40			&  0.2484  &  5.0838   &   8.3473	&  0.3215  &   0.746	&  0.882    & 0.931 \\
		\rowcolor{gray!9}
		Vankadari \textit{et al.} \cite{vankadari2020unsupervised}  &  M    &   40			&  0.2005 &  2.5750  &   7.172	&  0.278   &   0.735	& 0.883    & 0.942 \\
        Liu \textit{et al.} \cite{liu2021self} &  M    &   40			&  0.233 &  2.344  &  6.859	&  0.270   &   0.631	& 0.908    & 0.962 \\
		\rowcolor{gray!9}
		ITDFA$_{d2n}$(\textbf{JT})  &  M     &   40	&  0.2274 &  1.4327 &  5.5849 &  0.2855  &  0.573	&  0.889  & 0.953\\
		ITDFA$_{d2n}$(\textbf{ST})  &  M     &   40	&  \textbf{0.1469}  &  \textbf{0.9963} &   \textbf{4.6851} &  \textbf{0.2065}  &   \textbf{0.778}	&  \textbf{0.928}   & \textbf{0.973}\\%& \textbf{0.1656} &  \textbf{1.3902} &   \textbf{5.458}& \textbf{0.226}  &   \textbf{0.764}&   \textbf{0.912}   & \textbf{0.964} \\
		\hline
        %\multicolumn{10}{c}{Night-time scenario}\\
		\hline
		Monodepth2 (d) \cite{godard2019digging} 	&  M     &   60			&  0.5009  &  6.2895   &   11.5469	&  0.5100   &   0.295	&  0.538    & 0.756 \\
		\rowcolor{gray!9}
		Monodepth2 (n) \cite{godard2019digging} 	&  M     &   60			&  0.2899  &  6.8150  &   11.9479	&  0.3562   &   0.665	&  0.846    & 0.914 \\
		Vankadari \textit{et al.} \cite{vankadari2020unsupervised}  &  M    &   60			&  0.2327 &  3.783  &   10.089	&  0.319   &   0.668	& 0.844    & 0.924 \\
		\rowcolor{gray!9}
        Liu \textit{et al.} \cite{liu2021self} &  M    &   60	&  0.231 &  2.674  &  8.800	&  0.286   &   0.620	& \textbf{0.892}    & 0.956 \\
		ITDFA$_{d2n}$(\textbf{JT})  &  M     &  60	& 0.2789  & 2.4936 & 8.5216 & 0.3350  & 0.453	& 0.821   & 0.937\\
		\rowcolor{gray!9}
		ITDFA$_{d2n}$(\textbf{ST}) &  M     &   60	& \textbf{0.1869} &  \textbf{1.7752} &   \textbf{7.370}& \textbf{0.252}  &   \textbf{0.692}&   0.889  & \textbf{0.961}\\
		
		\bottomrule
		
	\end{tabular}
\end{table*}

\begin{table*}[h]
	
	\scriptsize
	
	\centering
	
	\caption{\textit{Comparison with the unsupervised depth estimation methods for rainy night-time scenarios. }}
	
	\label{Tab02}
	\begin{tabular}{c|c|c|cccc|ccc}
		
		\toprule
		\multicolumn{3}{c}{}& \multicolumn{4}{c}{Error Metrics (Lower is better)} & \multicolumn{3}{c}{Accuracy (Higher is better)}  \\
		\cmidrule(r){1-3}\cmidrule(r){4-7} \cmidrule(r){8-10}
		\hline
		Method   	&  Supervision     &   Depth-range (meter)		&  Abs Rel      &  Sq Rel    &   RMSE 	&  RMSE log     &   $\delta < 1.25^{1}$		&  $\delta < 1.25^{2}$      &  $\delta < 1.25^{3}$ \\
		\hline
		\rowcolor{gray!9}
		Monodepth2 (d) \cite{godard2019digging} 	&  M     &   40			&  0.4297  &  4.7547   &   8.780	&  0.463  &   0.369	&  0.640    & 0.828 \\
		Monodepth2 (r) \cite{godard2019digging} 	&  M     &   40			&  0.3902  &  50.8281  &   14.945	&  0.386  &  0.616	& 0.837   & 0.915\\
		\rowcolor{gray!9}
		ITDFA$_{d2r}$ &  M    &   40	&  \textbf{0.1642} &  \textbf{0.9688} &   \textbf{4.5737}	&  \textbf{0.2252}   &   \textbf{0.733}	& \textbf{0.932}   & \textbf{0.983} \\
		\hline
		ITDFA$_{d2n}$ 	&  M     &   40			&  0.1678  &  1.0283   &   4.6862	&  0.232  &   0.733	&  0.922    & 0.976 \\
		\rowcolor{gray!9}
		ITDFA$_{d2n2r}$ &  M     &   40	&  \textbf{0.1495} &  \textbf{0.9116}  &   \textbf{4.3658}	&  \textbf{0.2117}   &  \textbf{0.780}	& \textbf{0.940}    & \textbf{0.983} \\
		\hline
		\hline
		Monodepth2 (d) \cite{godard2019digging} 	&  M     &   60			&  0.4838  &  6.7168   &   11.357	&  0.509  &   0.343	&  0.596    & 0.781 \\
		\rowcolor{gray!9}
		Monodepth2 (r) \cite{godard2019digging} 	&  M     &   60			&  0.4211  &  48.4135  &   17.129	&  0.423  &  0.541	& 0.794   & 0.897 \\
		ITDFA$_{d2r}$ &  M     &   60	&  \textbf{0.1990} &  \textbf{1.7338} &   \textbf{6.9703}	&  \textbf{0.272}   &   \textbf{0.654}	& \textbf{0.882}   & \textbf{0.961} \\
		\hline
		\rowcolor{gray!9}
		ITDFA$_{d2n}$ 	&  M     &   60			&  0.2060  &  1.8000   &   7.0483	&  0.280  &   0.644	&  0.878    & 0.958 \\
		ITDFA$_{d2n2r}$ &  M     &   60	&  \textbf{0.1788} &  \textbf{1.5625}  &   \textbf{6.5729}	  & \textbf{0.2530} &  \textbf{0.716}   &   \textbf{0.902}	& \textbf{0.967}  \\
		
		\bottomrule
		
	\end{tabular}
\end{table*}

\begin{table*}[h]
	
	\scriptsize
	
	\centering
	
	\caption{\textit{Quantitative results for ablation study on RobotCar dataset \cite{RobotCarDatasetIJRR} using the night-time images. Depth range is 40m.}}
	
	\label{Tab03}
	\begin{tabular}{c|c|c|c|c|cccc|ccc}
		
		\toprule
		\multicolumn{2}{c}{$L_{FC}$}&\multicolumn{2}{c}{$L_{O}$}&\multicolumn{1}{c}{}& \multicolumn{4}{c}{Error Metrics (Lower is better)} & \multicolumn{3}{c}{Accuracy (Higher is better)}  \\
		\cmidrule(r){1-2}\cmidrule(r){3-4}\cmidrule(r){5-5}\cmidrule(r){5-9} \cmidrule(r){10-12}
		\hline
		$L_{FC_{L1}}$   	&  $L_{FC_{Gram}}$   &  $L_{OC}$   &  $L_{OS}$     &   Image adaptation &  Abs Rel  &  Sq Rel &   RMSE &  RMSE log &   $\delta < 1.25^{1}$	&  $\delta < 1.25^{2}$ &  $\delta < 1.25^{3}$ \\
        \hline
		$\surd$	&     & &     &  $\surd$	       &  0.1675  &  1.3213  &   5.2446 &  0.2342  &   0.749	&  0.908    & 0.962 \\
		\rowcolor{gray!19}
		$\surd$	&  $\surd$   & &     &  $\surd$	   &  0.1561  &  1.1678  &   4.9162 &  0.2161  &   0.767	&  0.921    & 0.967 \\
		&     & $\surd$ &    & $\surd$	           &  0.1637  &  1.1603  &   5.2037 &  0.2264  &   0.727	&  0.918   & 0.970 \\
		\rowcolor{gray!19}
		&     & $\surd$ &  $\surd $  & $ \surd	$  &  0.1664  &  1.0535  &   4.8925 &  0.2222  &   0.739	&  0.927    & 0.972 \\
		$\surd$	&   $\surd$  &$\surd$ &  $\surd$   &  $\surd$	&  \textbf{0.1469}  &  \textbf{0.9963} &   \textbf{4.6851} &  \textbf{0.2065}  &   \textbf{0.778}	&  \textbf{0.928}   & \textbf{0.973} \\
		\rowcolor{gray!19}
		$\surd$	&   $\surd $ &$\surd$ & $ \surd $  &  	        &  0.1545  &  1.0954 &   4.8664 &  0.2140  &   0.764	&  0.925    & 0.970 \\
		\hline
		\hline
		\multicolumn{1}{c}{$L_{FC}$} & \multicolumn{1}{c}{$L_{O}$} & \multicolumn{2}{c}{Image adaptation} &  ``N='' &  Abs Rel  &  Sq Rel &   RMSE &  RMSE log &   $\delta < 1.25^{1}$	&  $\delta < 1.25^{2}$ &  $\delta < 1.25^{3}$ \\
        \hline
		\rowcolor{gray!19}
		\rowcolor{gray!19}
		$\surd$	&   $\surd$  & \multicolumn{2}{c}{$\surd$}  &  latest	&  0.1501  &  1.0664 &   4.8121 &  0.2105  &   0.775	&  0.926   & 0.971 \\
		$\surd$	&   $\surd$  & \multicolumn{2}{c}{$\surd$}  &  50	&  0.1510  &  1.0585 &   4.7674 &  0.2118  &   0.775	&  0.926   & 0.970 \\
		$\surd$	&   $\surd$  & \multicolumn{2}{c}{$\surd$}  &  30	&  \textbf{0.1466}  &  \textbf{0.9824} &   \textbf{4.6649} &  \textbf{0.2062}  &   \textbf{0.780}	&  \textbf{0.929}   & \textbf{0.972} \\
		$\surd$	&   $\surd$  & \multicolumn{2}{c}{$\surd$}  &  20	&  0.1582  &  1.0953 &   4.8872 &  0.2174  &   0.750	&  0.922   & 0.971 \\
		
		\bottomrule
		
	\end{tabular}
\end{table*}

\begin{figure*}[t]
	\centering
	\subfigure[\textbf{Estimating the depth from night-time images.}]{
		\includegraphics[width = 0.95\columnwidth]{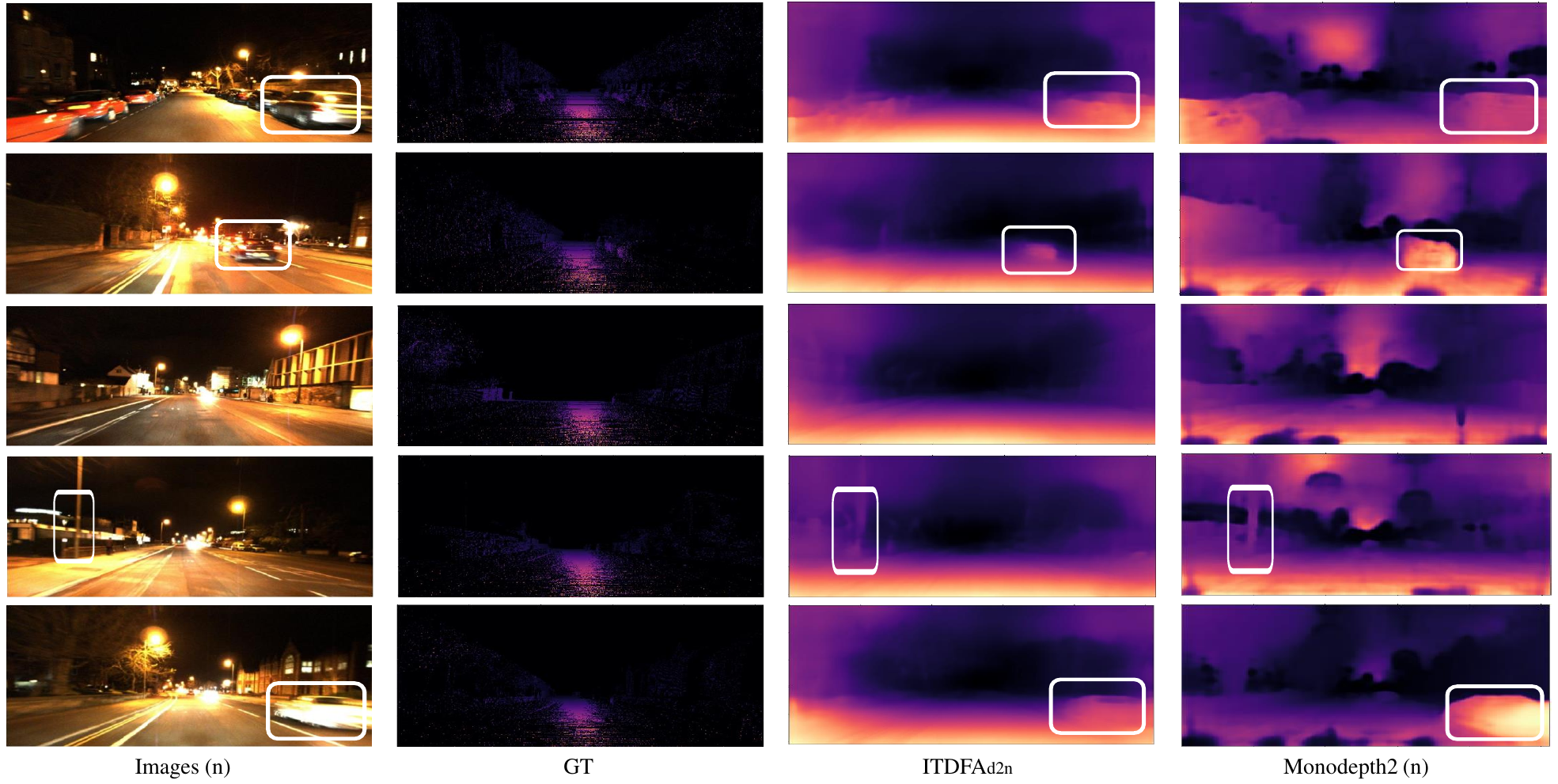}
		%\caption{fig1}
	}
	\subfigure[\textbf{Estimating the depth from rainy night-time images.}]{
		\includegraphics[width = 0.95\columnwidth]{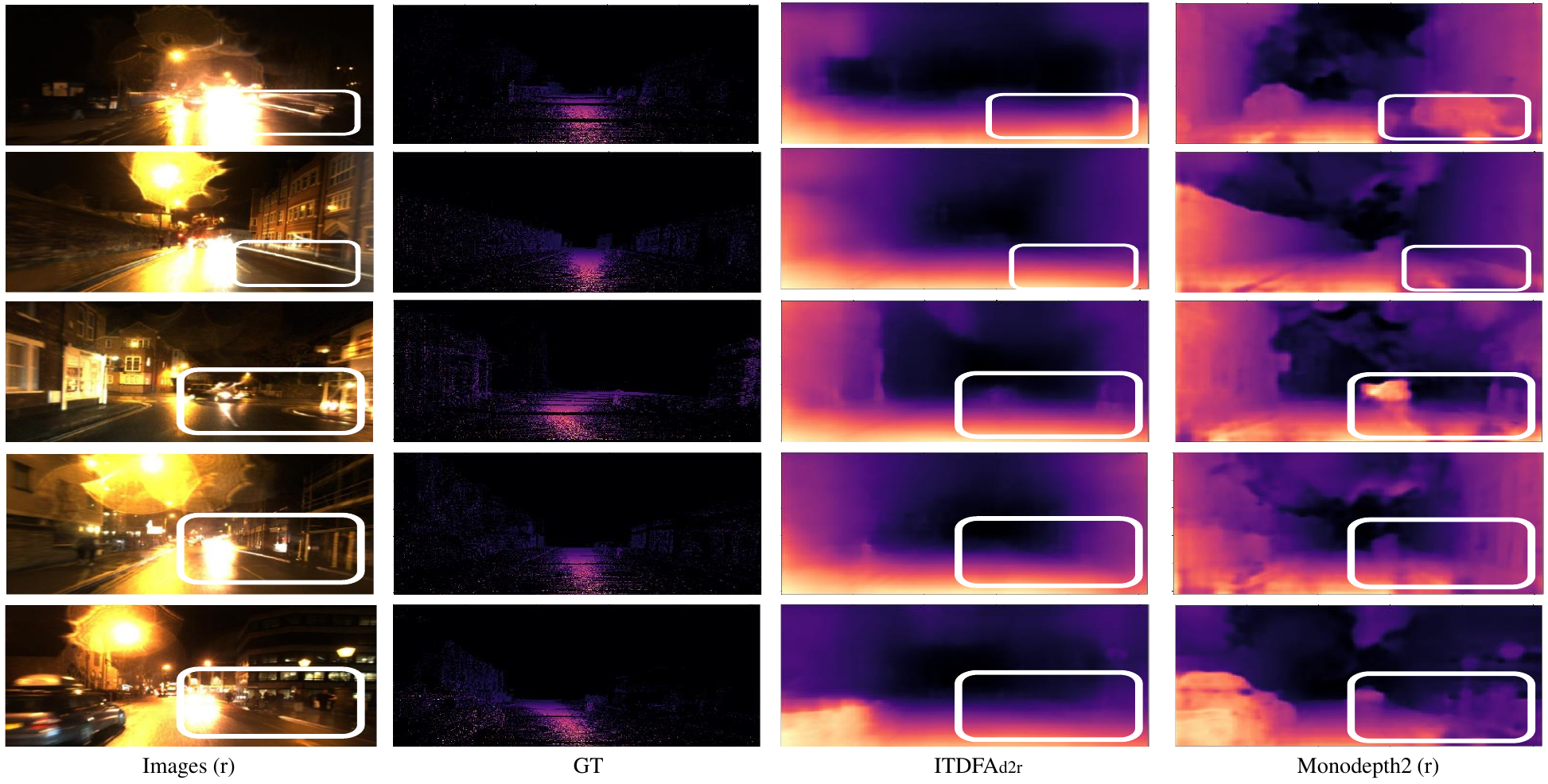}
	}
	\caption{In (a), ``Monodepth2(n)'' refers to the depth model trained on night-time images by the unsupervised monocular framework, monodepth2. For night-time scenario, it suffers from complex lighting conditions, like the street lights, car lights, and the reflection of light from the road, which bring big challenges to the unsupervised training framework. In (b), ``Monodepth2(r)'' refers to the depth model trained on rainy night-time images by the unsupervised monocular framework, monodepth2. For the rainy night scenario, it suffers from not only the complex lighting conditions at night but also the complex reflections on the road and camera caused by rain.}
	\label{fig:fig7}
\end{figure*}

\subsection{Training and testing sets setup}

\textbf{Training sets:} For the day-time depth model, the 5 splits of the day-time sequence are used to train the unsupervised framework \cite{godard2019digging}, and the basic pre-trained depth model, monodepth2 (day), is obtained for ITDFA. To improve the performance of this depth model, 15,000 images are uniformly selected from 5 splits for training, and the training set does not include the images taken while parking.
For the image transfer model, 5000 images of each scenario are random selected to obtain the image transfer models between different scenarios by using CycleGAN \cite{CycleGAN2017}. The selection of these images follows the rules before. To address the problem of unsupervised monocular depth estimation in night-time and rainy night-time scenarios, three image transfer models should be pre-trained for ITDFA: between day and night ($G_{d2n}$ and $G_{n2d}$), and between day and rainy night ($G_{d2r}$ and $G_{r2d}$). In addition, to verify the continuous transfer ability of the model obtained through domain adaptation, we also train an additional image transfer model between night and rainy night ($G_{n2r}$ and $G_{r2n}$).
During training the ITDFA, the training sets used for domain adaptation are the same as that of the image transfer model.

\textbf{Testing sets:} During testing, for the night scenario, the testing set is the same as  \cite{vankadari2020unsupervised} for a fair comparison, which contains 500 night images \footnote{\url{https://github.com/zxcqlf/RobotCar_DepthGT_Generate}}. While for the more complex rainy night scenario, 300 rainy night-time images are randomly selected from the remaining splits of each sequence to test the model obtained by ITDFA. The evaluation metrics used in this paper follow previous monocular methods \cite{godard2019digging,zhao2020monocular,vankadari2020unsupervised}, and we evaluate the depth models from the aspect of error and accuracy with different depth range (40m and 60m).

\subsection{Experimental setup}
The experiments are implemented by using Pytorch framework on an NVIDIA RTX 2080 Ti GPU. The network frameworks of depth models (including encoder and decoder) and image transfer model are the same with previous work monodepth2 \cite{godard2019digging} and CycleGAN \cite{CycleGAN2017}. To pre-train the depth model and image transfer model, we use the original setting proposed in monodepth2 \cite{godard2019digging} and CycleGAN \cite{CycleGAN2017} on the new datasets, and the resolution of images in training process is resized to 512x256. For the ITDFA, the framework is trained by using Adam optimizer \cite{kingma2014adam} with a batch size of 1. The learning rate is set as 0.0002 during training. The weights of the three loss components are set to $\alpha=15.0$, $\beta=0.01$ and $\gamma=0.001$. The experimental setup of ITDFA is the same when applied for different complex scenarios, including the night and rainy night. Moreover, the encoders trained for different complex scenarios share the same day-time decoder. Therefore, in practice, this framework only needs to switch the corresponding encoder to cope with the change of environments, which is practical and meaningful for applying in changing environments and all-day depth estimation.

As shown in Fig. \ref{fig:fig2} (b), the image transfer module and the depth adaptation module can adopt two modes: joint training (\textbf{JT}) and separate training (\textbf{ST}). For the \textbf{ST} mode, the image transfer module is trained in advance, and then the latest saved image transfer model is used to transfer the image between different scenario styles in depth adaptation. To get better cycle transfer models, we firstly test the image transfer models by transferring many sample images, and then we select those models without over-fitting in cycle transfer. According to the experiments, we use the transfer model saved at 30 epoch (N=30) to reflect the quality of the transferred images. For the \textbf{JT} mode, the image transfer module and the depth adaptation module are jointly trained. To get a more stable training process, we detach the gradient propagation between the two modules. Moreover, in the \textbf{JT} mode, we directly use the error between the reconstructed image and raw image to reflect the quality of the transferred image. ITDFA(\textbf{ST}) is trained for 50 epoches because the image transfer model is pretrained, while ITDFA(\textbf{JT}) is trained for 200 epoches.

\subsection{Results}

%\begin{figure}[t]
%	\centering
	%\includegraphics[width = \columnwidth]{figures/Depth.pdf}
	%\caption{A qualitative comparison of predicted depth-maps with different experiments. $X$ refers to the night-time scenario (lines 1, 2), rainy night-time scenario (lines 3, 4) and snowy winter scenario (lines 5, 6).%The first column includes the images from different complex environments $X$. ``Monodepth2(X)'' stands for the monodepth2 model trained and tested on the same scenario $X$. ``Monodepth2 (day)'' refers to the monodepth2 model trained on day-time scenario and tested on the complex scenario $X$. The final column shows the results of ITDFA for different complex environments $X$.
%}
%	\label{fig:fig4}
%\end{figure}

\subsubsection{\textbf{Annotation}}

Results of related experiments are shown in Table \ref{Tab01} and Table \ref{Tab02}. To the best of authors' knowledge, since the proposed work may be the first attempt at solving the monocular depth estimation problem in rainy night-time scenario, and no priors are available in the literature, we compare the results of different models based on the well known unsupervised framework, monodepth2 \cite{godard2019digging}.

In the tables, ``Monodepth2 (X)'' refers to the depth model trained on the images from `$X$' by the monodepth2 framework  \cite{godard2019digging}, and $X$ refers to day ($d$), night ($n$), rainy night ($r$). ITDFA$_{d2X}$ represents the depth model trained by the proposed ITDFA for the complex scenario `$X$', and these models (ITDFA$_{d2n}$, ITDFA$_{d2r}$) are adapted from the same day-time model, ``Monodepth2 (d)''. Moreover, ITDFA$_{d2n2r}$ stands for the rainy night-time model adapted from the night-time model, ITDFA$_{d2n}$, which means that the model is obtained through two domain adaptations by ITDFA. Note that all the models obtained by the proposed ITDFA framework share the same decoder with the ``Monodepth2 (d)''.  The qualitative results are shown in Fig. \ref{fig:fig7}. %ITDFA(\textbf{JT}) refers to the model trained by joint training mode, and ITDFA(\textbf{ST}) stands for the model trained by separate training mode.

\subsubsection{\textbf{Night-time depth estimation}}
To verify the effectiveness of the proposed ITDFA framework in night-time scenario, we compare our model with the state-of-the-art method \cite{vankadari2020unsupervised,liu2021self}, which only focuses on unsupervised night-time depth estimation. As shown in Table \ref{Tab01}, for estimating the depth from night-time images, the depth model trained by ITDFA gets a much better performance than the current state-of-the-art method \cite{vankadari2020unsupervised,liu2021self} with lower error and higher accuracy. Compared with the joint training mode ($ITDFA_{d2n}$(\textbf{JT})), the separate training mode ($ITDFA_{d2n}$(\textbf{ST})) gets more accurate adaptation results, because the pretrained image transfer model can provide more stable transferred images. Besides, the training time of the separate training mode for domain adaptation (1 day) is shorter than that of the joint training mode (1 week). Therefore, for the experiments of rainy night-time depth estimation, we adopt the separate training mode to train the depth models.
%Since their algorithms and trained models have not yet been published, we

\subsubsection{\textbf{Rainy night-time depth estimation}}

As shown in Table \ref{Tab02}, because of the domain drift, the monodepth2(d) does not have a good performance when testing on the rainy night-time images.
Compared with the day-time and night-time scenarios, the rainy night-time scenario suffers from not only photometric inconsistency but also the reflection of road and camera caused by rain. Therefore, the monodepth2(r) cannot get an accurate depth estimation because of the limitation of the unsupervised framework \cite{godard2019digging} in such complex scenario. The rainy night-time depth model trained by ITDFA (ITDFA$_{d2r}$) shows a much better performance than the monodepth2 models, which proves the effectiveness of the framework proposed in this paper.

\textbf{Double jump domain adaptation:} Because the domain gap between night and rainy night is smaller than that between day and rainy night, ITDFA$_{d2n}$ shows more accurate depth estimation than the monodepth2(d) on rainy night-time images.
To study the influence of different domain gaps on adaptation, we additionally design a two-step adaptation method, in which the day-time depth model is firstly adapted to night-time scenario, and then the night-time model is adapted to rainy night scenario, shown as ITDFA$_{d2n2r}$.
As shown in Table \ref{Tab02}, even if after the second round adaptation, ITDFA$_{d2n2r}$ achieves an outstanding accuracy than others in rainy night-time scenario, which means that multi-step adaptation will be helpful for the domain adaptation between large domain gap.
Besides, it also proves that the proposed ITDFA framework has the ability to effectively learn the key features for depth decoding, and these key features can be well adapted to new scenarios. The qualitative results are shown in Fig. \ref{fig:fig7}.%On the other hand, since the performance of the ITDFA is related to the image transfer model, and the performance of the image transfer model from night to rainy night is higher than the model from day to rainy night. Therefore, the model adapted from night-time models achieves the best performance for the rainy night-time depth estimation.

%\begin{figure}[t]
%	\centering
%	\includegraphics[width = 0.9\columnwidth]{figures/Fig7.pdf}
%	\caption{Different network frameworks are designed for comparison, including directly editing the features(1st line), training a novel encoder (2nd line, this paper) and training a full depth network (3rd line). Light orange modules indicate modules that need to be trained.}
%	\label{fig:fig10}
%\end{figure}

%\subsubsection{\textbf{Snowy winter depth estimation}}

%In the snow-covered environment during winter, the unsupervised framework suffers from \textit{repeat-texture} scenes, and the camera exposure and gain in outdoor snowy scenario affects the photometric consistency between frames, resulting in the limited performance of unsupervised framework. Therefore, as shown in Table \ref{Tab03}, the adaptive model obtained by ITDFA gets a better performance in snowy winter scenarios than the models trained by monodepth2. The qualitative results are shown in Fig. \ref{fig:fig4} and Fig. \ref{fig:fig9}, although the model ``Monodepth2'' looks better than our proposed method in details, the depth of some objects, like cars, is incorrect according to the relative depth. Meanwhile, due to different image brightness caused by the camera exposure and gain, the depth estimation of the similar road is significantly different, as shown in Fig. \ref{fig:fig4}.
%Quantitative results in Table \ref{Tab03} prove that our method can achieve good results in snowy winter. ITDFA achieves more robust depth estimation, but monodepth2 outperforms ITDFA in the depth presentation of details.

\subsubsection{\textbf{Ablation study}}
To analyze the effects of each component in the overall loss function $L_{total}$, Eq. (\ref{eq:7}), we design a series of ablations to analyze our approach, and quantitative results are shown in Table \ref{Tab03}.
Experiments show that the constraint of feature space is more effective than that of output space in promoting the consistency of feature maps, because in our framework, the night-time network shares the same decoder with the pretrained day-time models. Besides, the model trained by the constraints from both feature space and output space outperforms the others, which means that the constraints of output space help encoder to focus on learning the key features of depth decoding during training. The smoothness loss helps to improve the accuracy of depth estimation in complex environments. Moreover, the introduction of the proposed image adaptation method effectively improves the accuracy and reduces the error of monocular depth estimation.
%Compared with the model constrained by output space, the model constrained by feature space achieves higher performance, which means that the constraint of feature space is more effective to promote the consistency of feature space. While the model trained by the constraints from both feature space and output space outperform the others, so that . Moreover, the smoothness loss helps to improve the accuracy of depth estimation in complex environments.

\subsection{\textbf{Discussion}}
The ITDFA is an unsupervised framework, and the depth models for highly complex scenarios are trained in a completely unsupervised manner. Neither the image style transfer model nor the depth model use paired images or ground truth labels during training process.
Note that all the monocular depth models trained by ITDFA for different environments share the same decoder during testing, which has practical significance. For example, in autonomous driving, facing different weather conditions \cite{chu2017camera}, the vehicle can independently switch to the corresponding encoder to obtain better environmental perception.
Although this paper focuses on the effects of unstable transfer on image transfer-based domain adaptation, the proposed method cannot completely eliminate the effects, and the choices of transfer and cycle transfer models are important for the performance of the overall adaptation framework.

\section{Conclusion}

In this paper, we tackle the problem of unsupervised monocular depth estimation in highly complex environments, which is important and practical for autonomous systems. We survey the related research on solving the limitations of current unsupervised monocular depth estimation framework, and analyze the reason why the unsupervised framework cannot do well in certain highly complex environments. A novel domain adaptation framework, called ITDFA, is proposed in this paper to address the above problem. The proposed ITDFA framework is totally unsupervised and does not use any ground truth labels in the training process. Our method considers the shortcomings of image transfer-based domain adaptation approach and achieves more accurate depth estimation in night-time scenario than the state-of-the-art \cite{vankadari2020unsupervised,liu2021self}. Moreover, the performance of ITDFA in the more challenging rainy-time scenario proves the practicability and effectiveness of ITDFA. Therefore, ITDFA is able to provide a way to address the complex environmental change problems faced by monocular depth estimation during practical application. Furthermore, there are still shortcomings that need to be addressed, like enhancing the depth perception of some small objects in complex environments, which is also a promising direction for future work.

% or


\ifCLASSOPTIONcaptionsoff
  \newpage
\fi

% number - used to balance the columns on the last
{%\small
	\bibliographystyle{IEEEtran}
	\bibliography{egbib}
}


%\vfill


\end{document}